\documentclass[5p,times, authoryear]{elsarticle} % 3p
\usepackage{xurl}

\makeatletter
\def\ps@pprintTitle{%
 \let\@oddhead\@empty
 \let\@evenhead\@empty
 \def\@oddfoot{\centerline{\thepage}}% or \def\@oddfoot{} %
 \let\@evenfoot\@oddfoot}
\makeatother

\usepackage[utf8]{inputenc}
\usepackage[switch]{lineno}

\journal{Journal of Cachexia, Sarcopenia and Muscle}

\title{Comprehensive Validation of Automated Whole Body Skeletal Muscle, Adipose Tissue, and Bone Segmentation from 3D CT images for Body Composition Analysis: Towards Extended Body Composition}
\author[1,2]{Da Ma}
\author[1,2]{Vincent Chow}
\author[1,2]{Karteek Popuri}
\author[1,2]{Mirza Faisal Beg \corref{cor1}}

\address[1]{Simon Fraser University, School of Engineering Science, Burnaby, British Columbia, Canada}
\address[2]{Voronoi Health Analytics Incorporated, Vancouver, British Columbia, Canada}

\cortext[cor1]{Correspondence address: research@voronoihealthanalytics.com, Voronoi Health Analytics Inc., Vancouver, British Columbia, Canada}

\begin{document}

\begin{frontmatter}

\begin{abstract}

% == Background/Introduction
\textbf{Background}
The latest advances in computer-assisted precision medicine are making it feasible to move from population-wide models that are useful to discover aggregate patterns that hold for group-based analysis to patient-specific models that can drive patient-specific decisions with regard to treatment choices, and predictions of outcomes of treatment. Body Composition is recognized as an important driver and risk factor for a wide variety of diseases, as well as a predictor of individual patient-specific clinical outcomes to treatment choices or surgical interventions. 3D CT images are routinely acquired in the oncological worklows and  deliver accurate rendering of internal anatomy and therefore can be used opportunistically to assess the amount of skeletal muscle and adipose tissue compartments. Powerful tools of artificial intelligence such as deep learning are making it feasible now to segment the entire 3D image and generate accurate measurements of all internal anatomy. These will enable the overcoming of the severe bottleneck that existed previously, namely, the need for manual segmentation, which was prohibitive to scale to the hundreds of 2D axial slices that made up a 3D volumetric image. Automated tools such as presented here will now enable harvesting whole-body measurements from 3D CT or MRI images, leading to a new era of discovery of the drivers of various diseases based on individual tissue, organ volume, shape, and functional status. These measurements were hitherto unavailable thereby limiting the field to a very small and limited subset. These discoveries and the potential to perform individual image segmentation with high speed and accuracy are likely to lead to the incorporation of these 3D measures into individual specific treatment planning models related to nutrition, aging, chemotoxicity, surgery and survival after the onset of a major disease such as cancer. 

\textbf{Methods}
We developed a whole-body 3D CT segmentation method for body composition analysis. This method delivers accurate 3D segmentation of the bony tissue, the skeletal muscle, subcutaneous and visceral fat. 

% == Results
\textbf{Results}
Evaluation on 50 volumes in the evaluation dataset achieved an average dice coefficients of 0.980 for bone, 0.974 for skeletal muscle, 0.986 for SAT, and 0.960 for VAT, demonstrating the accurate performance of the proposed whole-body 3D tissue segmentation algorithm.

% == Conclusion
\textbf{Conclusion}
The validation presented in this paper confirms the feasibility of going beyond 2D single-slice based tissue area measurements to full 3D measurements of volumes and texture for these tissues as well as other internal organ volumes, shape, location, texture and radiomic texture features in the body. Recent advances in machine/deep learning are capable of not only providing these measurements, but can also provide the necessary statistical models to learn the importance of each of these measurements in developing individual specific models of patient outcomes. This is a necessary step to unleash the full power of AI into supporting new and personalized treatment approaches in the domain of precision medicine.
\end{abstract}


\end{frontmatter}

\section{Introduction}

%% ========== Backgrounds
The study of body composition, namely the study of skeletal muscle (SKM), subcutaneous (SAT), visceral (VAT) and intramuscular (IMAT) adipose tissue, has been shown to drive clinical outcomes (toxicity, survival, infection, surgery, length of hospital stay). Specifically, SKM is an important prognostic factor for various outcomes such as cancer survival \cite{Penna2019} and surgical outcomes \cite{Gomibuchi2020,Nishimura2019}. For identical height and weight, SKM mass can vary substantially and for those outcomes that depend on skeletal mass (such as chemotoxicity), accurate measurement of the individual's SKM is very critically important. Therefore, measurements based on height and weight such as BMI or BSA are shown to not provide the accuracy needed to estimate the relative volumes of tissue compartments required to build models for predicting individual clinical outcomes. 

% ==========   CT images
CT images can provide accurate visualization of internal anatomy, and are routinely-acquired in oncology.
Estimation of body composition measures from 2D single-slice analysis (L3/T4) is now widely established \citep{Dabiri2019,Dong2020,Paris2020}. The field of body composition has been driven mainly by analysis of single-slice-based 2D measures from routinely acquired CT images in the oncologic workflow, where it was shown that these 2D measures correlate to 3D volumetric measurements \citep{Shen2004a}.
However, although single slice area measures such as on L3 can be a proxy for whole body measures in group studies, it is questionable whether 2D single-slice measures are accurate for predicting individual clinical outcomes. Indeed, in the original study \citep{Shen2004a}, it was shown that single-slice-derived metrics could only reliably distinguish subjects differing in skeletal muscle volume by at least 10.4 liters and adipose tissue volume by at least 12.4 liters giving an error range of 10-30\%; indicating limited accuracy in correctly estimating whole-body muscle and fat volumes in an individual subject and therefore of limited utility in building individual specific dosing, toxicity or survival prediction models. This study also showed that different axial slices provided highest correlation to whole body measures for skeletal muscle and fat, whereas over time, the field settled on using the same slice for all tissues. Further more, the regression equation estimated in this study may not generalize to all age ranges, or diseased groups, and the angle of the axial slice can also vary based on the position of the individual with respect to the scanning axis. based prediction may not be accurate for individual. When performing group studies, these errors may cancel out leading to reasonably accurate final results, but when the goal is individual prediction and treatment decisions, such as dose of chemotherapy, it is imperative that all possible sources of error be minimized.

On the other hand, there is an amazing wealth of information available in 3D CT images on which the single-slice approaches have been based, with the limiting factor being the ability to efficiently analyze and extract measurements from all the hundreds of slices that can make a patients' individual scan. Many critical measurements, such as volumes and functional status of internal organs, such as the liver, cannot be performed in a single-slice manner, and were therefore hitherto excluded from such analyses. Also, topographic distribution of tissues, such as abdominal cavity fat versus mediastinal cavity fat could be important for prognosis, and this delineation is unavailable in single slice analysis. The main bottleneck of harvesting the wealth of information present in CT images can now be overcome thanks to the availability of platforms such as convolutional neural networks. Therefore, the next generation of models can and should move beyond single-slice measures of SKM/VAT/SAT/IMAT to fully 3D measurements of these tissues, ideally whole body measurements, and not be limited to these a-priori assumptions that only SKM/VAT/SAT/IMAT tissues contribute to patient specific outcomes. The expanded repertoire of measurements from individualized tissues and organs is likely to enable the discovery of new biomarkers of aging, sarcopenia, obesity and cachexia, in particular, “subtypes” of muscle loss and fat gain in aging and disease and contribute to greater understanding and better treatment design for the individual.

\section{Methods}

\subsection{Ground truth data preparation and curation}
Ground truth segmentation labels were manually generated by a team of trained anatomists using the semi-automatic segmentation tools using an in-house manual segmentation platform. Four tissue types were generated: 1) skeletal muscle (SKM) 2) bone; 3) subcutaneous adipose tissue (SAT), and 4) the visceral adipose tissue (VAT). On average, it took two weeks for each trained anatomist to carefully delineate these tissues manually for each volume highlighting the enormous time and skill that such measurements require.

\subsection{Evaluation of segmentation performance}
The performance of our proprietary algorithm was evaluated on a set of 50 volumes that were manually segmented.  Dice similarity coefficient (Equation \ref{eq:dice}) was used to evaluate the the segmentation accuracy for each tissue type.
\begin{equation}
    \frac{2(A\bigcap B)}{|A|+|B|}
    \label{eq:dice}
\end{equation}
where A is the ground truth label maps generated from manual segmentation, and B is the automatically-generated label maps generated from the proposed whole-body 3D segmentation framework.

\subsection{Population wide variation of whole body tissue distribution}
% proof of concept
A pilot proof-of-concept study on the variation of the SKM, bone, VAT and SAT tissues as well as their average attenuation (HU) measures across a database of males and females was conducted to demonstrate the increased analytical capability of whole body 3D tissue segmentation. We manually identified the vertebrae from the segmented bone labels. A total of 21 vertebral segmentations were extracted: 4 from cervical vertebra (C4-C7), 12 from the thoracic vertebra (T1-T12), and 5 from lumbar vertebra (L1-L5), and labelled. For each vertebral segmentation, we extracted the corresponding slab of the volumes containing the axial slices that it belongs to. The total tissue volumes, and the mean HU value for each tissue types were calculated for each vertebra-indexed slabs.

\section{Results}
In this section, we first present the results of our whole-body 3D segmentation framework both qualitatively and quantitatively. We also present the analysis result for the intra-subject and cross-population variation of the whole body distribution of body composition, indexed by the location of the vertebrae along the axial slice direction, both in terms of tissue volume as well as the tissue attenuation level.

\subsection{Qualitative visual evaluation of segmentation performance}
% \subsubsection{Individual tissue segmentation}

\begin{figure*}
    \centering
    \includegraphics[width=16cm]{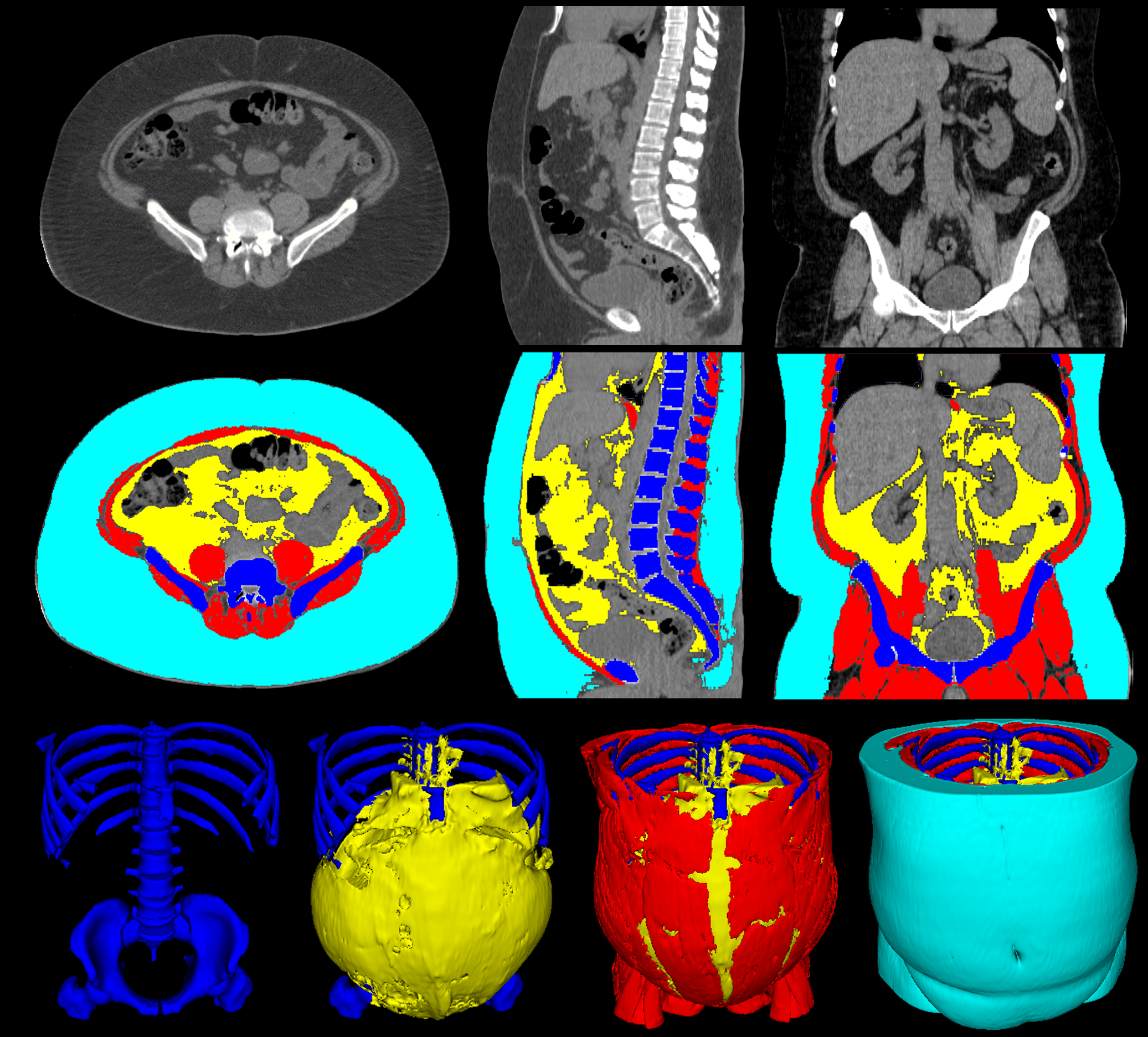}
    \caption{Representative image of the  automated 3D whole-body segmentation of tissues from a 3D volumetric CT scan.
    (Top): The three-plane view (axial, sagittal, and coronal) of original 3D CT volumetric scan; 
    (Middle): The three-plane view (axial, sagittal, and coronal) of the automatic tissue segmentation
    of four tissues namely the bone (blue), skeletal muscle (red), VAT (yellow), and SAT (cyan).
    (Bottom): 3D renderings of the segmented tissue, from left-to-right: bone, VAT overlaid on the bone to show its topographic distribution, muscle, VAT and bone, and finally, SAT overlaid on the others showcasing a full 3D volumetric segmentation.
    }
    \label{fig:sample_segmentation}
\end{figure*}

Figure \ref{fig:sample_segmentation} is a representative images of a sample full-body automated 3D segmentation of four tissue types: skeletal muscle, bone, SAT, and VAT, demonstrating accurate tissue boundaries. The corresponding 3D rendering of each tissue types (Figure \ref{fig:sample_segmentation} bottom row) provides an effective way to visualise each tissue compartment for each individual scan. For this individual scan, the axial, coronal and sagittal slices show a marked presence of SAT, whereas the 3D renderings in the bottom row show a marked presence of VAT compared to a relative paucity of skeletal muscle.

\begin{figure*}
    \centering
    \includegraphics[width=16cm]{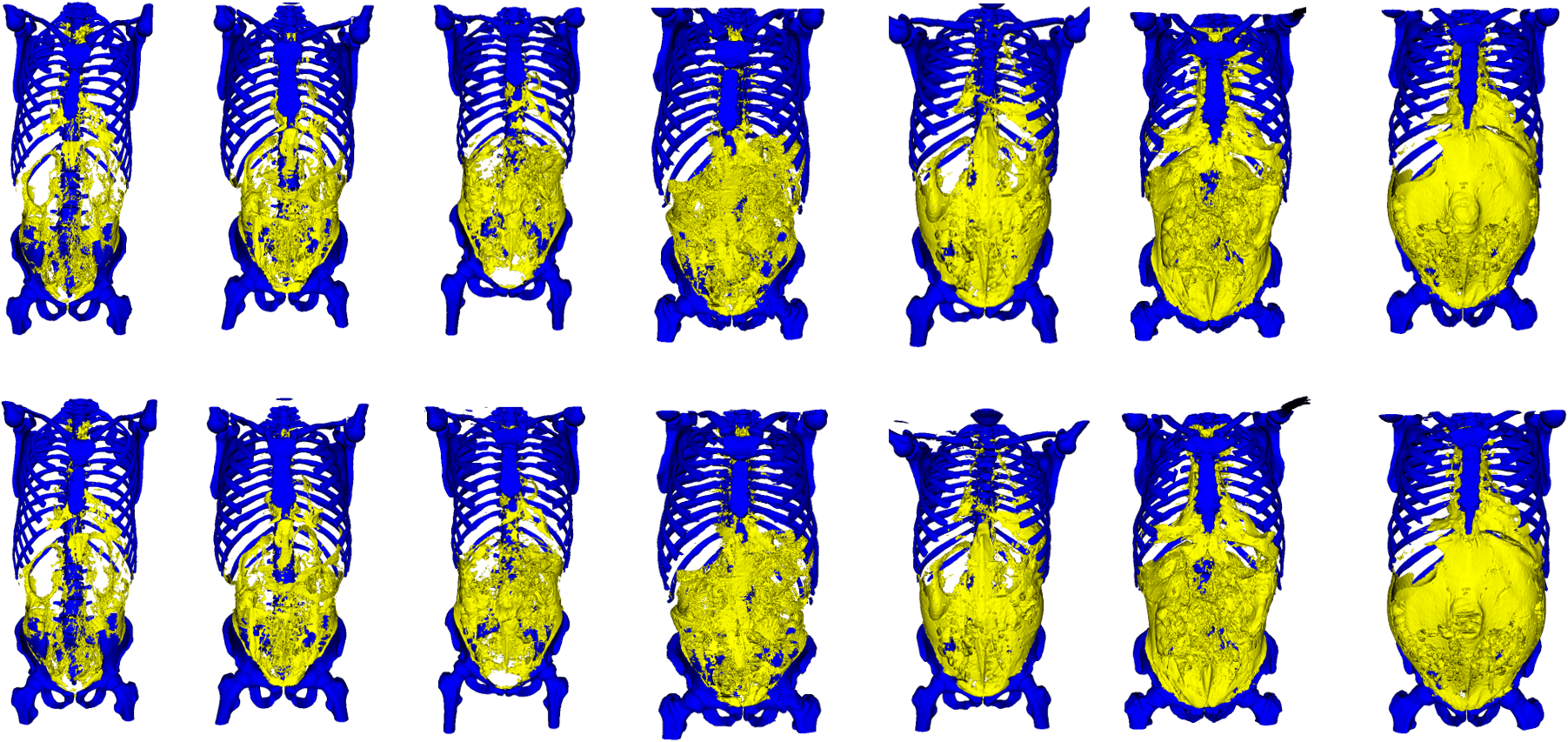}\\
    \includegraphics[width=16cm]{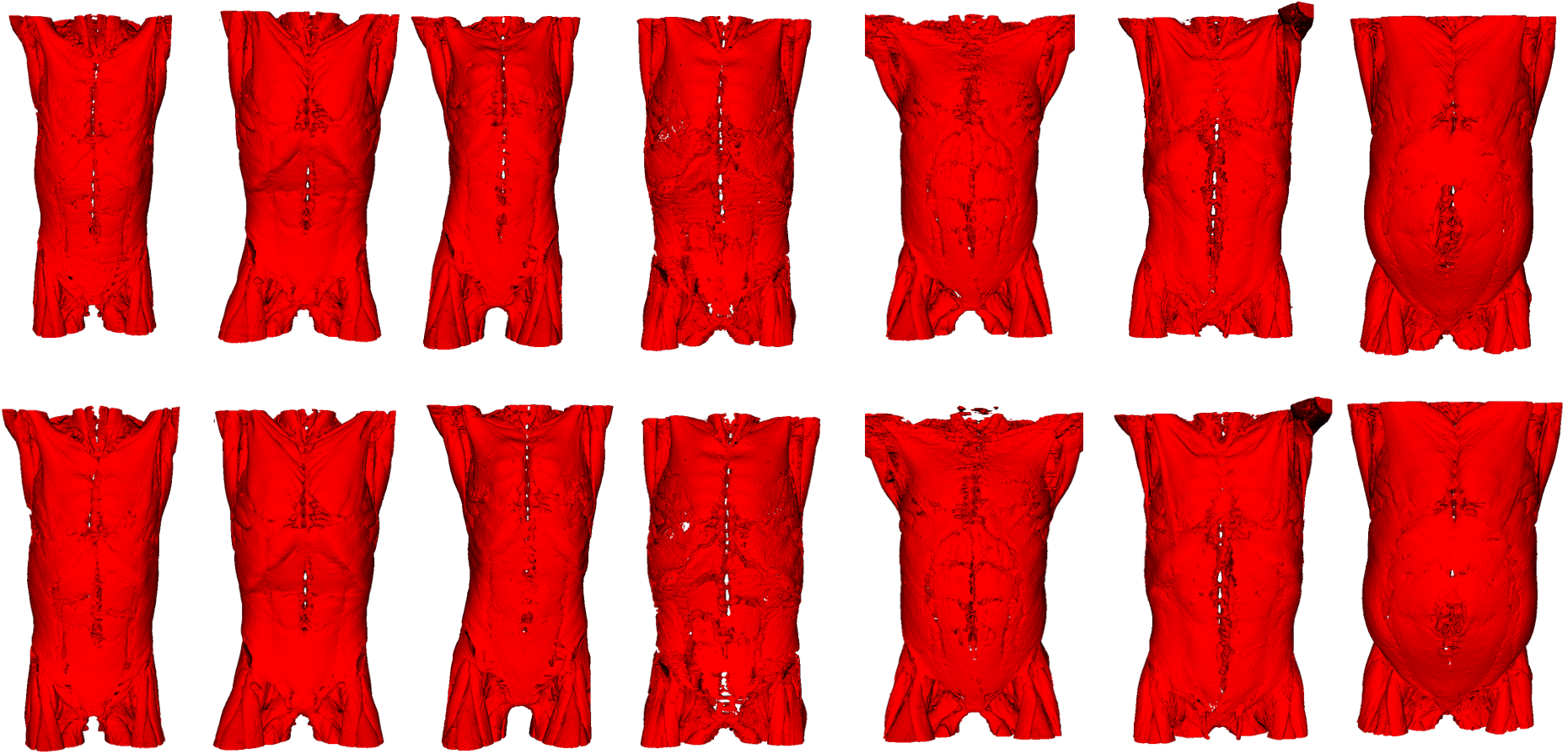}
    \caption{Visual comparison between manual and auto segmentation for muscle (in red), fat (in yellow) and bone (in blue).
    1st row: manual segmentation label map of whole body bone and visceral adipose tissue (VAT).
    2nd row: automatic segmentation label map of whole body bone and VAT.
    3rd row: manual segmentation label map of whole body skeletal muscle (SKM).
    4th row: automatic segmentation label map of whole body SKM.
    }
    \label{fig:sample_manual_vs_auto}
\end{figure*}

Figure \ref{fig:sample_manual_vs_auto} demonstrates the visual comparison between the ground truth manual tissue labels maps (1st and 3rd row) and the automated segmented tissue labels map (2nd and 4th rows) side-by-side, demonstrating an almost perfect agreement between the two. Furthermore, comparison of the 3D visualization among multiple subjects (across rows) showed not only overall intra-subject body composition variation, but also variation in terms of tissue distribution within each individual. Increasing volume of VAT can be observed in subjects from left to right.

\subsection{Quantitative evaluation of segmentation performance}

\begin{figure}[htb!]
    \centering
    \includegraphics[width=9cm]{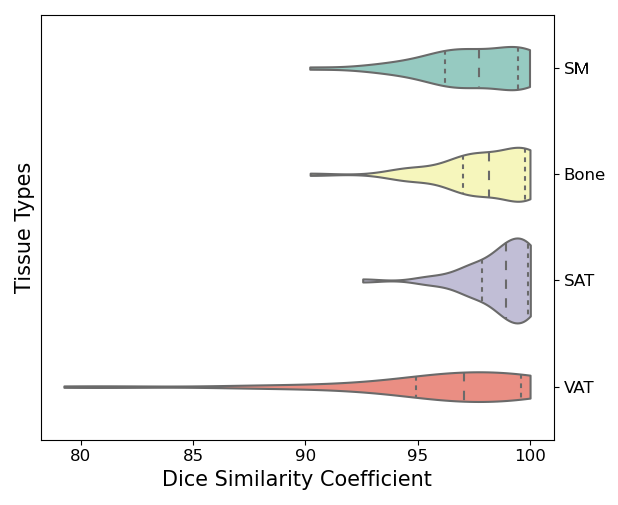}
    \caption{Violin plot of the Dice coefficient for each of the four segmented tissues. Note the high Dice coefficient with a larger fraction of the volumes positioned towards the higher end indicating good performance and potential for generalization. }
    \label{fig:dice_violin_plot}
\end{figure}

We evaluated the segmentation performance of the proposed model on the 50 volumes in the evaluation dataset. The average dice coefficients achieved on this test set are found to be 0.980 for bone, 0.974 for skeletal muscle, 0.986 for SAT, and 0.960 for VAT, demonstrating the accurate performance of the proposed whole-body 3D tissue segmentation algorithm.
Figure \ref{fig:dice_violin_plot} shows the corresponding violin plot, both the mean and standard deviation statistics as well as the population distribution of the Dice similarity coefficient for each segmented tissue on the previously unseen evaluation dataset.

\subsection{Analysis of cross-population variation of whole-body distribution of body composition}

\subsubsection{Cross-body tissue volumetric distribution}
% do we need to perform statistical analysis on these distribution curves? (probably not necessary for the Arxiv version, but might be necessary for the journal version)

In addition to measuring the overall segmentation accuracy with the Dice coefficient, we also evaluated the variability of measures of body composition  provided by the proposed 3D whole-body tissue segmentation on a subset of the population. Specifically, the 3D rendering qualitative results shown in Figure \ref{fig:sample_manual_vs_auto} demonstrated the individual variation of the tissue distributions across the entire body, revealing features that are beyond the capability of traditional 2D single-slice measures. Figure \ref{fig:body_distribution} showed further quantitative analysis of the proposed 3D whole body composition analysis, in terms of the distribution of the measured tissue volume (top) and average attenuation value (HU) across the entire sagittal direction of the whole body, indexed by the vertebrate level. As a proof-of-concept analysis, we also plotted the whole body tissue distributions for male and female subjects separately, and compared the sex-dependent differences.

% different distribution among different tissue
When using vertebrae locations to index the tissue measures, tissue-specific whole body volume distribution patterns can be observed (Figure \ref{fig:body_distribution} top row). Specifically, the accumulation of VAT in the range of the L1-L5 is highly variable across individuals, while the distribution of SAT is more uniform. 
% sex-dependent volume difference
Interestingly, when comparing the whole body volume distribution of VAT, a noticeable difference could be observed between the male (blue curves) and female (red curves) sub-population, where female subjects typically showed lower accumulation of VAT in the lumbar-index regions of the body

\subsubsection{Cross-sectional tissue CT attenuation distribution}

\begin{figure*} % [htb]
    \centering
    \includegraphics[width=16cm]{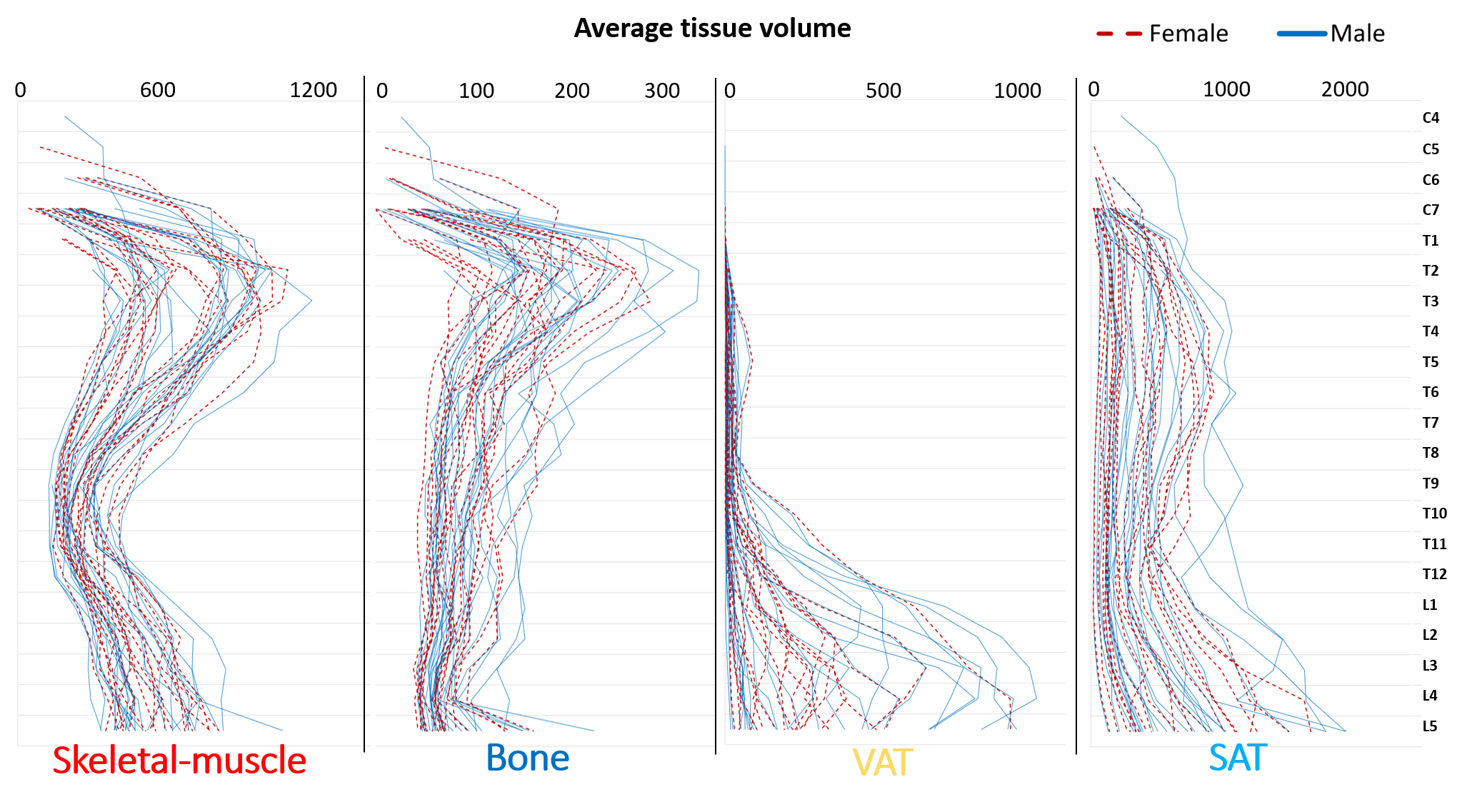}\\
    \includegraphics[width=16cm]{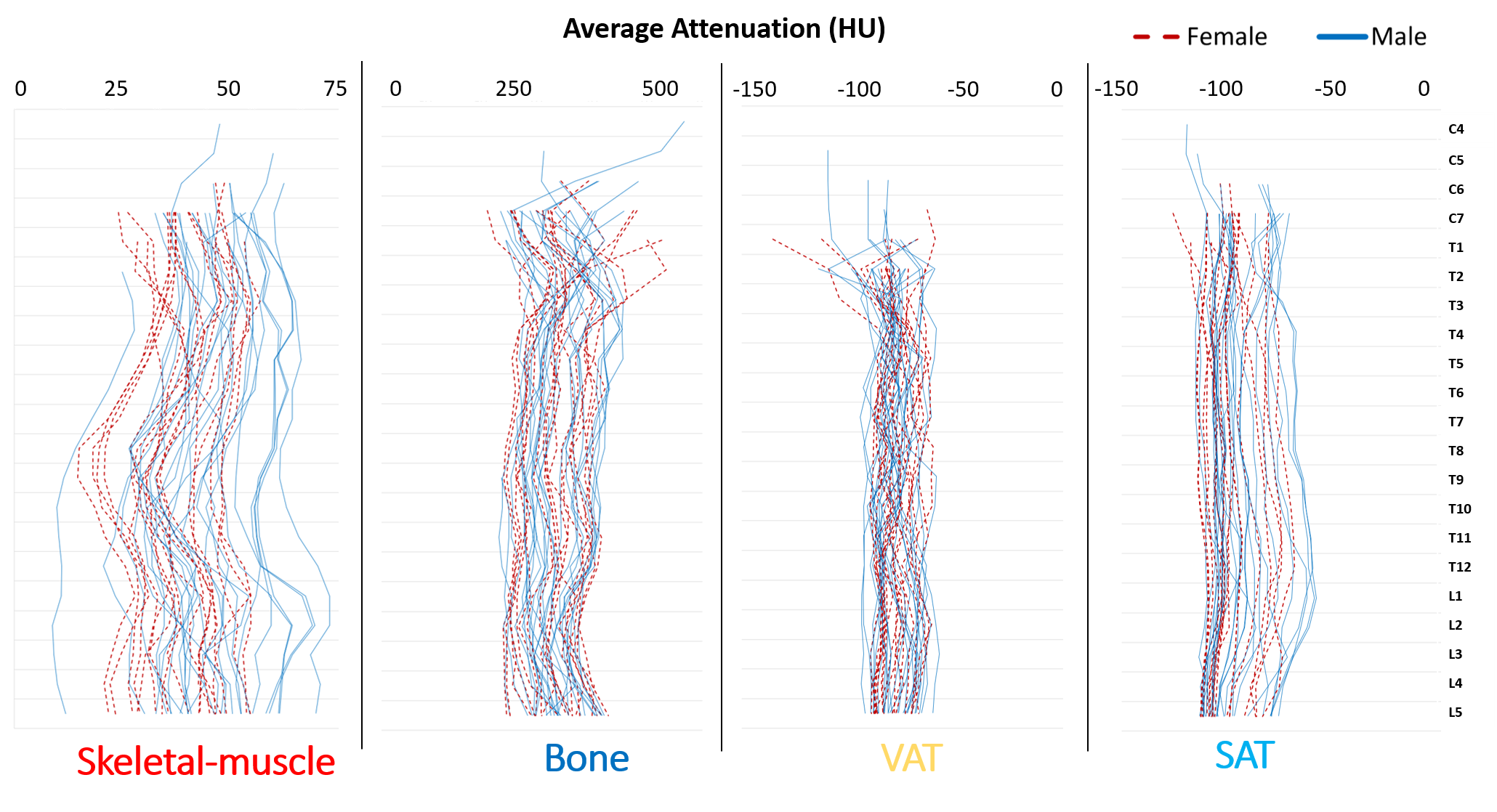}
    \caption{Distribution of (Top) tissue volumes (cm$^3$) and (Bottom) CT average attenuation value (HU) for each tissue type, indexed by vertebral level plotted separately for female (in dashed red line) and male (in solid blue line). Two skeletal muscle phenotypes are apparent in the volumetric distribution, with marked variance at the T2-T3 level along with considerable variation in the VAT and SAT across individuals topographically. These topographic signatures may have useful predictive ability in addition to whole body volumetric measures.
}
    \label{fig:body_distribution}
\end{figure*}

%%%%%%% HU distribution
Compared to the tissue volume, less variation was observed for the distribution of attenuation level (indicated by the HU value) across different vertebrate-index body regions (Figure \ref{fig:body_distribution} bottom row), with the exception of skeletal-muscle, indicating fluctuations in the muscle properties across the body. The attenuation level appear to be higher for the muscles that reside in the top thoracic-indexed axial slices (T1-T8) and top lumbar-indexed axial slices (T1-T3).
In addition, the pattern of distribution of attenuation measures is more homogeneous across sex. However, the overall attenuation level appear to be higher for the male population compared to the female population.
For the adipose tissues, a sex-dependent attenuation difference appeared in the SAT similar to that for the muscle, although with lower range of difference. However, interestingly, no obvious such sex-dependent attenuation different could be observed for the VAT.

% I feel there is a lack of statistical analysis to support all the above observed statements. Yes! We need to make these more precise! But next version ... lets get this into arxiv...

\section{Conclusion and Discussion}
% Conclusion
In this study, we have presented the construction and evaluation of our 3D tissue segmentation whole body composition framework. Evaluation of the framework on 50 3D CT volumes showed excellent agreement compared to the manually delineated ground truth tissue segmentation. The vertebrae-indexed whole body distribution of the tissue volume and attenuation value showed distinctive distribution patterns among different tissue types, and indicated a certain level of sex-dependence in some of these measures, demonstrating the power of 3D whole-body body composition analysis.

\subsection{Comparison between 2D single-slice and 3D volumetric BC analysis}
%%  limitation of 2D single slice analysis
When conducting body composition analysis, one important question to ask would be: is 2D single-slice BC analysis accurate enough for predicting individual clinical outcomes? In the original 2D analysis paper by \citep{Shen2004a}, it was mentioned that, same 2D slice may not be best for both skeletal muscle (SKM) and adipose tissue (AT), with optimal axial slice located at 5cm above L4-L5 for skeletal muscle, while 5cm below L4-L5 for the adipose tissue. Such inconsistency between tissue types in terms of optimal location for 2D to 3D correlation adds errors when measures from a single slice are used for both SKM and fat as proxy for their distribution across the body. In addition, \cite{Shen2004b} also showed gender-based differences in terms of the optimal slice location that have the highest correlation between the 2D area and 3D volume for VAT: 10 cm above L4-L5 in men and 5cm above L4-L5 in women, further indicating the limitations of 2D single-slice based analysis. Currently, the field has settled on the L3 axial slice as the slice of choice to analyze for measures of body composition. However, as mentioned by \cite{Shen2004a}, the regression-based approach may be adequate for group studies but was not suitable for individual decision making since the regression-based predictions of whole body measures could not distinguish individuals accurately. This relevant text from \cite{Shen2004a} is quoted here because it is highly important: 
\begin{quote}
\textit{
Because single-slice imaging can only be used to distinguish reliably between subjects who differ in SM volume by $\ge 10.4–12.0$ liters and AT volume by $\ge 12.3–15.1$ liters, the use of a single-image SM or AT area estimate, therefore, has limited application when individual subjects are evaluated. However, in group studies, using an appropriate single-slice area would require only 17$\%$ more subjects for estimating total body SM and 6$\%$ more subjects for estimating total body AT than a multi-slice, whole body volume protocol. Using a single-slice area at L4-L5 would require 24$\%$ more subjects for SM and 12$\%$ more subjects for AT estimates than a multi-slice, whole body volume protocol, indicating that an appropriately selected slice can be used for group studies. For future large-scale studies, investigators thus have the option of increasing the subject sample size and thereby reducing the complexity and cost of image acquisition and analysis. Nevertheless, because increasing the sample size may be difficult and costly for certain kinds of studies (e.g. subjects with diseases), a whole body imaging protocol might remain the best choice.
}
\end{quote}

More importantly, the correlation analysis in previous studies were measured from relatively young subjects which may not generalize to other cohorts such as the elderly, sarcopenic, obese, or different disease groups where significant alterations in body composition may be present. Body composition measures from a single-slice also can introduce uncertainty and errors due to the angle of the single-slice leading to the cut-induced variability that is dependent on the variability of the tissue compartment. This would lead to unavoidable errors specifically in analyzing individuals over time, even if the same vertebral location was selected, as the axial slice acquired could be at a different angle. The highest sensitivity to slice angle is likely observed in the intramuscular fat and visceral fat compartments, which can change dramatically from slice to slice and as a function of the angle of the axial slice. For the purposes of cross-sectional population group analysis, such errors may cancel out, but when these measures are used for individual prediction, they may be inaccurate. Further, the drivers of disease onset and progression could be the topographic distribution of tissues, and the relative distribution of the tissues across the body, and these measures are unavailable in the single-slice analysis approach.

The 3D distribution results in Figure \ref{fig:body_distribution} show that performing measurements in 3D are feasible and reveal a great wealth of information on the topographic distribution of the individual tissue compartments. Seemingly uniform tissues (such as fat) are known to be  heterogeneous, where the biology, gene expression and behavior of adipocytes are dependent on their location. By analyzing fat tissues in 3D, such location-dependence can be incorporated in the analysis, with future versions of body composition measuring fat tissues in the ventral, abdominal and pelvic cavities and further subdividing ventral fat into mediastinal fat etc. enabling a much more fine-grained view.  The separation of the pericardial adipose tissue (PAT) and subcutaneous adipose tissue (SAT), the 3D PAT/SAT ratio \cite{Alman2017} and SAT/VAT ratio \cite{Mittal2019} can also be measured accurately, providing further clinical relevant information.
Further, the tissue remodeling dynamics could be the result of a complex interplay between location and function, with core muscles more preserved in those who remain ambulatory as compared to those who are immobile, and some individuals may have a phenotype of depositing fat into SAT compartments versus VAT compartment. Going to 3D will also enable the sub-segmentation of the full extent of individual muscles to help assess localized functional remodeling in individualized muscles/tissues. 

\subsection{The concept of Extended Body Composition}

\begin{figure*}[h]
    \centering
    \includegraphics[width=18cm]{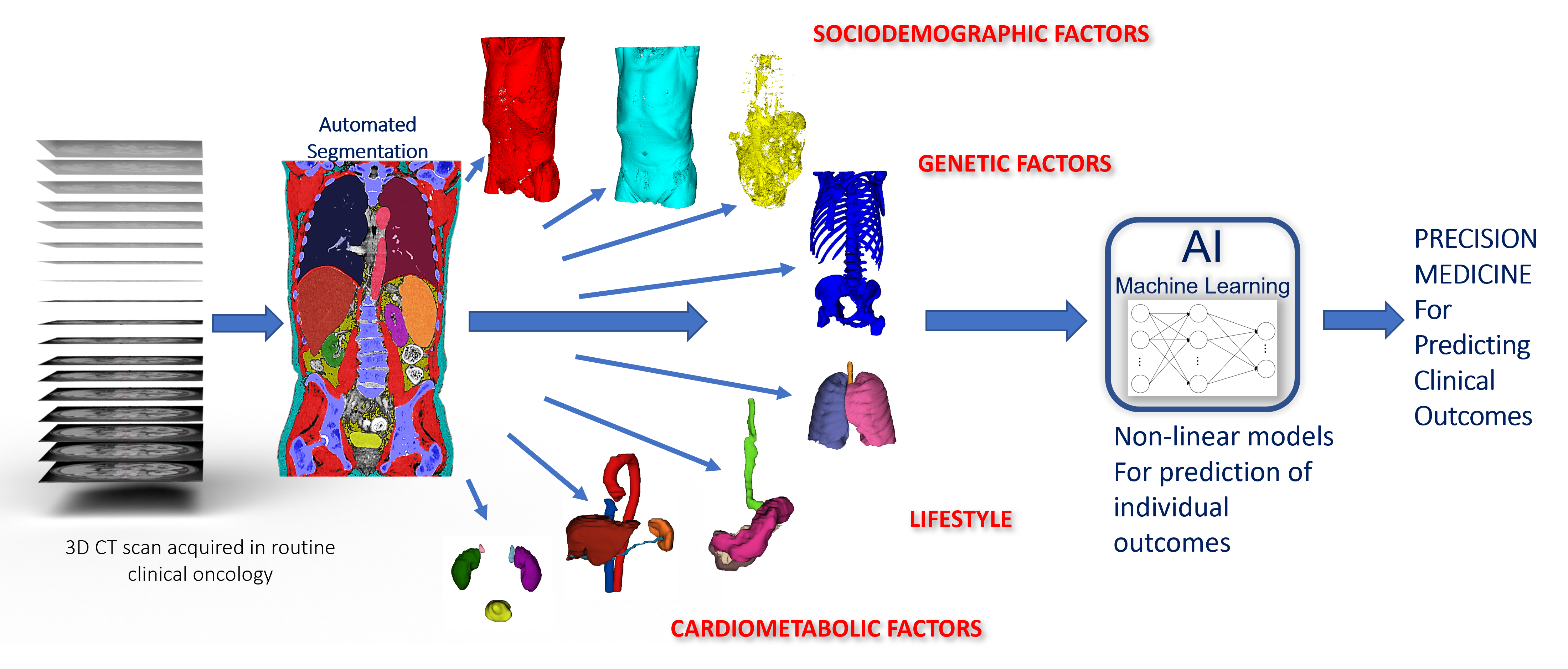}
    \caption{The concept of "Extended Body Composition": There is tremendous potential in harvesting the full extent of information regarding tissues and organs present in the 3D CT images. Going beyond the single-slice measures of SKM, VAT and SAT and incorporating volumes, shape, location, texture and radiomic texture analysis of other tissues and organs in what is termed "extended body composition" could enable the discovery of novel phenotypes and drivers of diseases and new treatment design strategies tailored for individual phenotypes. }
    \label{fig:future_vision}
\end{figure*}

Going beyond the traditional single slice measures to 3D measures of SKM/VAT/Sat is the first important step in enabling these measurements to be performed with reduced noise allowing their use in precision medicine. Further, 3D CT images provide a wealth of information that could be critically important in determining the individual phenotype as driver of certain risk factors, as well as allowing more fine-grained selection of treatment strategies for the individual going beyond "group-analysis". The volumes, tissue attenuation, and radiomic texture measures of other organs such the liver (the site of hundreds of physiological pathways), the spleen, the kidneys, the pancreas and other organs in the body are central to the "extended body composition" approach that is now feasible with the advent of AI-based segmentation approaches as conceptualized in Figure \ref{fig:future_vision}. These organs are highly variable and distributed across multiple slices, and therefore their analysis cannot be performed from single slice measurements, necessitating accurate 3D segmentation. This would enable discovering, in a data-driven manner, which variables are important and thereby lead to highly accurate patient-specific models for predicting clinical outcomes (chemo-toxicity, surgery, survival).

This would also cement the important role of 3D medical imaging in providing individual tissue delineation over other techniques such as electrical impedance and dual energy X rays (DXA) which can only provide aggregate tissue measures. Future extension of AI-based models can also integrate automated analysis of PET imaging (function/metabolism), genomics (role of individual genetic predisposition), markers of bone health (cortical and trabecular measurements), heart health markers (such as coronary and aortic calcium deposition), as well as nutritional status and individual lifestyle to enable the discovery of a new class of drivers for disease and biomarkers of disease onset and progression. Regression-based models are useful proxies for 3D measurements, but have inherent limitations in precision medicine where individual prediction is needed. Group studies are important in discovering results that hold on average for the group, but ultimately, the clinical and lifestyle interventions need to optimal for the individual. For enabling truly customized precision medicine incorporating these measures of extended body composition, it is imperative that direct 3D measurements available in the individual images be performed to overcome the unavoidable errors inherent in the single-slice approach that relies on regression and correlations as proxies to the true 3D measurements.

\subsection{Limitation of the current study and future work}
In the current study, the proposed 3D whole body segmentation framework is evaluated on a set of 50 CT scans (Figure \ref{fig:body_distribution}). In-field testing of this method by several researchers on unseen (to the classifier during training) CT images including both contrast and non-contrast images has provided empirical evidence that the algorithm as developed does generalize very well, however, further quantitative evaluation on larger and more diverse population will be necessary to estimate the errors that may be present, specifically in a test-retest setting and longitudinally, over time, from multiple scans acquired for an individual.  The on-going and future research directions involve questions around how to handle the heterogeneity in the field of view that may be present in retrospectively acquired CT scans, the development of models for partial and variable fields of view, and the covariates necessary to harmonize the extended body composition measurements accounting for stature differences. In addition, more generalizable models are required to facilitate robustness to CT acquisition parameters (contrast, non-contrast, low dose CT) and artifacts. Furthermore, automated segmentations need to also segment images containing idiosyncratic and/or unusual features such as fluid retention that is sometimes present in edematous patients, ascites, peritonial fluid or other idiosyncratic anatomy. 

The 3D segmentation framework described in this paper is commercially available as part of the Data Analysis Facilitation Suite 3.0 software platform, which can be obtained from \url{https://www.voronoihealthanalytics.com/dafs3}.  The availability of such AI-based algorithms as presented in this work are enabling a new approach to precision medicine that was not hitherto possible, and therefore holds tremendous promise for precision medicine. Finally, it is possible to construct longitudinal patient-specific quantitative models to deliver accurate assessment of localized changes from baseline to followup from 3D CT scans. This would enable the effect of prescribed clinical or lifestyle interventions and thereby the selection of targeted interventions that are highly optimal for the individual for achieving the promise of precision medicine.

\subsection{Acknowledgment}
This paper is based on concepts presented at the NIH Body Composition and Cancer Outcomes Research Webinar Series on December 17th, 2020 by Mirza Faisal Beg titled "Automating Body Composition from Routinely Acquired CT images - towards 3D measurements". The talk is archived at \url{https://epi.grants.cancer.gov/events/body-composition}

\bibliographystyle{plainnat}
\bibliography{main}

\end{document}